%% file: main.tex
\documentclass[runningheads]{llncs}
\pdfoutput=1 

\usepackage[utf8]{inputenc}
\usepackage{hyperref}
\usepackage[misc]{ifsym}
\usepackage{graphicx}
\usepackage{wrapfig}
\usepackage{verbatim}
\usepackage{paralist,amsmath,color,enumerate,framed}
\usepackage{todonotes}
\newcommand{\protege}{Prot\'eg\'e}
\newcounter{cqcounter}
\title{Towards a Modular Ontology for Space Weather Research}
\author{Cogan Shimizu\inst{1}\orcidID{0000-0003-4283-8701}\Letter \and Ryan McGranaghan\inst{2}\orcidID{0000-0002-9605-0007} \and Aaron Eberhart\inst{1}\orcidID{0000-0003-3007-5460} \and Adam C. Kellerman\inst{3}\orcidID{0000-0002-2315-936X}}
\institute{Data Semantics Laboratory, Kansas State University, USA \and Atmosphere and Space Technology Research Associates (ASTRA LLC), USA \and Department of Earth, Planetary, and Space Sciences, University of California, Los Angeles, USA}
\authorrunning{Shimizu, C., McGranaghan R., Eberhart, A., and Kellerman, A. C.}
\begin{document}
\maketitle
\input{abstract}
\input{sections/introduction}
\input{sections/example}

\input{sections/pattern}
\input{sections/conclusion}
\paragraph{Acknowledgements.} The authors acknowledge partial support from the National Science Foundation under Grant Numbers 1936677 and 1937152, as well as thorough discussion with Sean Gordon, Lisa Kerr, and Philip Murphy.
\bibliographystyle{splncs04}
\bibliography{refs}
\end{document}

%% file: abstract.tex
\begin{abstract}
    The interactions between the Sun, interplanetary space, near Earth space environment, the Earth's surface, and the power grid are, perhaps unsurprisingly, very complicated. 
    The study of such requires the collaboration between many different organizations spanning the public and private sectors. Thus, an important component of studying space weather is the integration and analysis of heterogeneous information.
    As such, we have developed a modular ontology to drive the core of the data integration and serve the needs of a highly interdisciplinary community. 
    This paper presents our preliminary modular ontology, for space weather research, as well as demonstrate a method for adaptation to a particular use-case, through the use of existential rules and explicit typing. 
\end{abstract}

%% file: sections/introduction.tex
\section{Introduction}
\label{sec:intro}
The solar-terrestrial system is complex, and consists of many interconnected subsystems nonlinearly interacting; energy flows from the surface of the Sun, through interplanetary space, into the near Earth space environment, and across the surface of the Earth. The expanse of the system as well as resource limitations dictate that a heterogeneous collection of data and models must be seamlessly utilized together \cite{McGranaghan_2017}. 

Interactions across the solar-terrestrial system are manifest on Earth in myriad ways that pose threats to our technological infrastructure and ability to use the near-Earth space environment \cite{Schrijver_2015}. Among the most important, yet ironically, least well specified impacts is that of the electric power grid. During periods of enhanced space weather activity, a series of physical processes beginning with a solar event (e.g., the launch of a coronal mass ejection (CME) or a high speed stream (HSS) from the Sun) gives rise to intense electric currents reaching millions of Amperes surrounding the Earth, which then become electric currents on the ground flowing through electrical transmission lines. This phenomenon, known as Geomagnetically Induced Currents (GICs), can disrupt the operation of high-voltage power grid transformers via overheating and generation of harmonics, potentially leading to failures \cite{Pulkkinen_2017}.

Power grid utility operators must stitch together observational and simulation data to know the appropriate responses to be taken. This is challenged by the number of disciplines that must cooperate and the intricate linking of the data across those disciplines involved--from space scientists, to earth scientists, to power system owners and operators, and power consumers.

What is needed is a semantic understanding of the notion linking solar events to power grid responses to form the foundation of an effective data integration system to study space weather and protect the power grid \cite{Narock_2012}. 

This paper describes the first modular ontological pattern that links the Sun to the power grid, specifically to guide decisions for power grid utilities. 


%% file: sections/example.tex
\section{Use-case Scenario}
\label{sec:use}
The Convergence Hub for the Exploration of Space Science (CHESS) project aims to unify the traditionally disparate communities and data sets that span the Sun, interplanetary space, near Earth space environment, the surface of the Earth, and the power grid. GICs arise from a series of interactions across these domains, beginning with the solar cloud of plasma interacting with the Earth's magnetic field, creating currents in space and in the upper atmospheric region known as the ionosphere, which produces the electric field on the ground through magnetic induction. However, knowledge of many aspects of this chain is limited However, knowledge of many aspects of this chain is limited, largely due to the interdisciplinary nature of each interconnected link. Therefore, the motivating question is, ``What is the ontological notion that links energy generated at the Sun, propagated through interplanetary space, and efficacious of disturbance to the power grid?'' Important in this question is the responses that each link requires (e.g., conducting a predictive simulation, monitoring real-time data streams, or taking mitigative action to protect the grid) such that the ontological design must connect with ODPs for simulation activities, sensor networks, and the actions that must be taken by power grid utility operators. We detail the ontology designed for CHESS, demonstrating the value of the modular design approach to facilitate interoperability and reusabiliity of existing ontological patterns.

%% file: sections/pattern.tex
\section{The Preliminary Ontology}
\label{sec:pattern}
This section first introduces the Data Transformation ontology design pattern and its instantiation, the Simulation Activity Module, which are key pieces of the ontology. 
We then show how the module fits into the modular ontology for space weather. 
Next, we describe a mechanism for adapting the domain ontology to a particular use-case and, 
finally, provide a worked example.

The OWL files for the pattern, module, and ontology can be found online, as well as our supplementary data.\footnote{See \url{https://github.com/cogan-shimizu-wsu/DataTransformationPattern}.} The OWL files were generated using the CoModIDE plugin \cite{comodide} and annotated with OPLa \cite{opla,opla-cp} using the OPLa Annotator plugin \cite{oplatool}; both plugins are for \protege{}\footnote{See \url{https://protege.stanford.edu/}.} 

\subsection{The Data Transformation Pattern}
In many fields, the transformation of data from one form to another, whether that be data-sanitation, analysis, or formatting, is a first-class operation. The Data Transformation pattern, herein described, allows for the description of data-driven workflows.

\begin{figure}[tb]
    \centering
    \includegraphics[width=\textwidth]{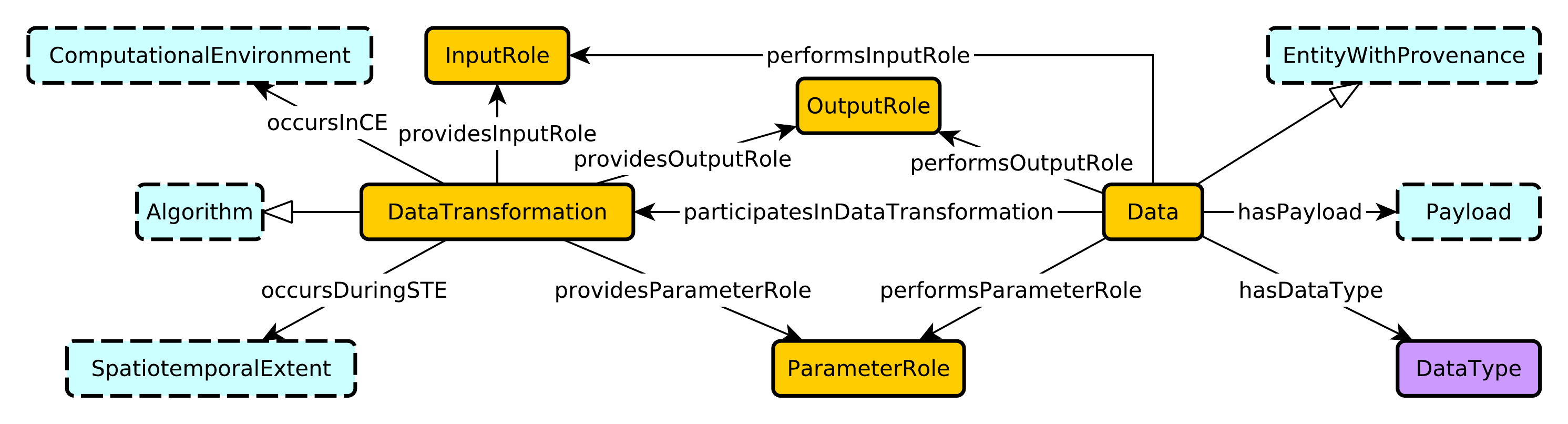}
    \caption{The schema diagram for the \textsf{DataTransformation} pattern. Yellow boxes denote classes. Blue boxes with dashed outline denote external patterns or modules. Purple boxes denote controlled vocabularies. Solid arrows denote object or date properties, while open arrows denote the subclass relationship.}
    \label{fig:pattern}
\end{figure}

In January 2020, as part of a joint venture for the NSF's Convergence Accelerator: Open Knowledge Networks initiative, several teams met to discuss the data modeling needs for each of the teams. It quickly became apparent that tracking data workflows, as well as availability of data, and even how that data might be used at all, was a very important aspect to each team's use-case, as we see in Section \ref{sec:use}. During this meeting we identified a number of interesting competency questions that serve to further frame the goals; a selection follows, with the entire document being found in the supplementary data.
\begin{enumerate}[CQ1.]
    \item What datasets are available to view?
    \item What does dataset \emph{X} contain?
    \item In what ways is dataset \emph{X} used?
    \item What is the result of dataset \emph{X} transformed by Algorithm \emph{A}?
    \item What dataset \emph{X} was used for input to Simulation \emph{S}?
    \setcounter{cqcounter}{\value{enumi}}
\end{enumerate}
The resultant pattern can be shown in Figure \ref{fig:pattern}. The core concept is \textsf{DataTransformation}. A \textsf{DataTransformation} always occurs in space and time (\textsf{Spatiotemporal\-Extent}) and in some \textsf{ComputationalEnvironment}. Each one also provides a number of roles: \textsf{InputRole}, \textsf{OutputRole}, and \textsf{ParameterRole}. These roles allow us to disconnect the \textsf{Data} from the transformation, acknowledging that the same data set can be used in different datasets and that no transformation is truly destructive. However, we do say that each instance of a role (as a transformation may have many, say, input data) may only be performed by exactly one \textsf{Data}. Further, every \textsf{Data} is an \textsf{EntityWithProvenance}, meaning that it is generally known from where data is sourced. We also specify, using the Explicit Typing pattern \cite{modl}, that \textsf{Data} might have a \textsf{DataType}, which allows us to specify, in a domain dependent manner, broad categories of data or format of data.\footnote{The explicit declaration of types allows us to easily change domain dependent information, without impacting an ontology's subsumption hierarchy. We posit that this strategy can aid with downstream alignment and co-reference resolution.}

We see in the schema diagram that \textsf{Algorithm}, \textsf{ComputationalEnvironment}, and \textsf{EntityWithProvenance} are represented by blue dashed boxes. This means that they are external to the pattern. That is, this pattern acknowledges that these concepts exist, and are integral to the notion of a \textsf{DataTransformation}, but makes no claims on their ontological structure. For the latter two, we recommend the use of their eponymous counterparts in \cite{modl}. \textsf{Algorithm} is very domain dependent and we leave it to be specified in specific use-cases.

\subsubsection{Axiomatization} This axiomatization was produced using the out-of-the-box axioms depicted in the Edge Inspector tool in CoModIDE \cite{comodide}, as well as the same axioms used during the systematic axiomatization process particular to Modular Ontology Engineering \cite{moe-chap}. They are provided in description logic syntax below. The axioms are segmented to aid reading. Axioms 2-10 are scoped ranges (e.g. axiom 2) and existentials (e.g. axiom 3). Axiom 1 is a subclass. Axiom 11 is a \emph{structural tautology} which serves to inform users of the intended use of an axiom. Axioms 12-16 and 29-32 are scoped domain axioms. 

Each of the properties included in this pattern have scoped domain and range, as we want to limit the ontological commitments made by this pattern. Each of the ``performs'' properties also have inverse existential and functionality. This means that only exactly one \textsf{data} can perform a role provided by a \textsf{DataTransformation}.

\begin{figure}[tb]
    \begin{align}
        \textsf{DataTransformation} &\sqsubseteq \textsf{Algorithm}\\
        \textsf{DataTransformation} &\sqsubseteq \forall\textsf{occursInCE.ComputationalEnvironment}\\
        \textsf{DataTransformation} &\sqsubseteq \exists\textsf{occursInCE.ComputationalEnvironment}\\
        \textsf{DataTransformation} &\sqsubseteq \forall\textsf{occursDuringSTE.SpatiotemporalExtent}\\
        \textsf{DataTransformation} &\sqsubseteq \exists\textsf{occursDuringSTE.SpatiotemporalExtent}\\
        \textsf{DataTransformation} &\sqsubseteq \forall\textsf{providesInputRole.InputRole}\\
        \textsf{DataTransformation} &\sqsubseteq \exists\textsf{providesInputRole.InputRole}\\
        \textsf{DataTransformation} &\sqsubseteq \forall\textsf{providesOutputRole}\\
        \textsf{DataTransformation} &\sqsubseteq \exists\textsf{providesOutputRole}\\
        \textsf{DataTransformation} &\sqsubseteq \forall\textsf{providesParamaterRole.ParameterRole}\\
        \textsf{DataTransformation} &\sqsubseteq \mathord{\geq}0\textsf{providesParamaterRole.ParameterRole}
    \end{align}
\end{figure}

\begin{figure}[tb]
    \begin{align}
        \exists\textsf{occursInCE.ComputationalEnvironment} &\sqsubseteq \textsf{DataTransformation}\\
        \exists\textsf{occursDuringSTE.SpatiotemporalExtent} &\sqsubseteq \textsf{DataTransformation}\\
        \exists\textsf{providesInputRole.InputRole} &\sqsubseteq \textsf{DataTransformation}\\
        \exists\textsf{providesOutputRole.OutputRole} &\sqsubseteq \textsf{DataTransformation}
    \end{align}
\end{figure}

\begin{figure}[tb]
    \begin{align}
        \textsf{Data} &\sqsubseteq \textsf{EntityWithProvenance}\\
        \textsf{Data} &\sqsubseteq \forall\textsf{performsInputRole.InputRole}\\
        \textsf{Data} &\sqsubseteq \exists\textsf{performsInputRole$^-$.InputRole}\\
        \textsf{Data} &\sqsubseteq \mathord{\leq}1\textsf{performsInputRole$^-$.InputRole}\\
        \textsf{Data} &\sqsubseteq \forall\textsf{performsOutputRole.OutputRole}\\
        \textsf{Data} &\sqsubseteq \exists\textsf{performsOutputRole$^-$.OutputRole}\\
        \textsf{Data} &\sqsubseteq \mathord{\leq}1\textsf{performsOutputRole$^-$.OutputRole}\\
        \textsf{Data} &\sqsubseteq \forall\textsf{performsParameterRole.ParameterRole}\\
        \textsf{Data} &\sqsubseteq \exists\textsf{performsParameterRole$^-$.ParameterRole}\\
        \textsf{Data} &\sqsubseteq \mathord{\leq}1\textsf{performsParameterRole$^-$.ParameterRole}\\
        \textsf{Data} &\sqsubseteq \forall\textsf{participatesInDataTransformation.DataTransformation}\\
        \textsf{Data} &\sqsubseteq \forall\textsf{hasPayload.Payload}\\
        \textsf{Data} &\sqsubseteq \forall\textsf{hasDataType.DataType}
    \end{align}
\end{figure}

\begin{figure}[tb]
    \begin{align}
        \exists\textsf{performsParameterRole.ParameterRole} &\sqsubseteq \textsf{Data}\\
        \exists\textsf{participatesInDataTransformation.DataTransformation} &\sqsubseteq \textsf{Data}\\
        \exists\textsf{hasPayload.Payload} &\sqsubseteq \textsf{Data}\\
        \exists\textsf{hasDataType.DataType} &\sqsubseteq \textsf{Data}
    \end{align}
\end{figure}

\subsection{The Simulation Activity Module}
\label{ssec:module}
\begin{figure}[tb]
    \centering
    \includegraphics[width=\textwidth]{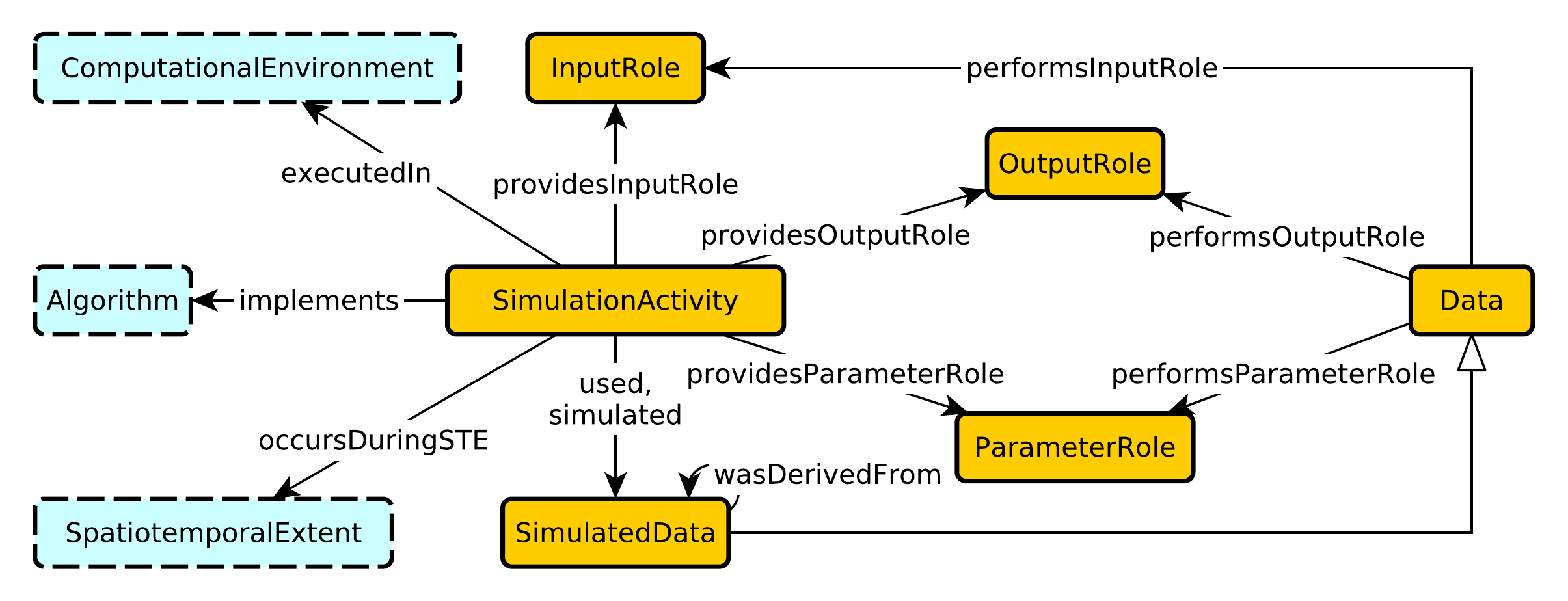}
    \caption{The schema diagram for the \textsf{SimulationActivity} module. It utilizes the same notation as previous diagrams.}
    \label{fig:module}
\end{figure}

The Data Transformation pattern is a necessarily general concept. During the ontology development process, we adapted the pattern to the use-case outlined in Sections \ref{sec:intro} and \ref{sec:use}. To do so, we made use of the strategy outlined in \cite{template}, which is called ``Template-based Instantiation.'' That is, we generally replaced names of classes and properties, instead of using subclass relationships. However, to keep the provenance of the pattern's structure, we use annotations from the Ontology Design Pattern Representation Language (OPLa) to indicate the original pattern. Figure \ref{fig:module} shows the new module.

The major addition is concept of \textsf{SimulatedData} and additional specification of its own provenance. These changes to the pattern allow us to answer the following competency questions that are more specific to the use-case.

\begin{compactenum}[CQ1.]
    \setcounter{enumi}{\value{cqcounter}}
    \item What were the solar wind conditions at the time of a given GIC observation \emph{G}?
    \item Is this GIC node at risk of exceeding a threshold GIC level in the next \emph{X} time?
    \item What are the simulated conditions in given the results of a simulation?
\end{compactenum}

The axioms for the additions are as follows.
\begin{align}
    \textsf{SimulationActivity} &\sqsubseteq \forall\textsf{used.SimulatedData}\\
    \textsf{SimulationActivity} &\sqsubseteq \forall\textsf{simulated.SimulatedData}\\
    \textsf{SimulatedData} &\sqsubseteq \forall\textsf{wasDereivedFrom.SimulatedData}
\end{align}


\subsection{The Modular Ontology for Space Weather}
\label{ssec:ontology}
\begin{figure}[tbp]
    \centering
    \includegraphics[width=\textwidth]{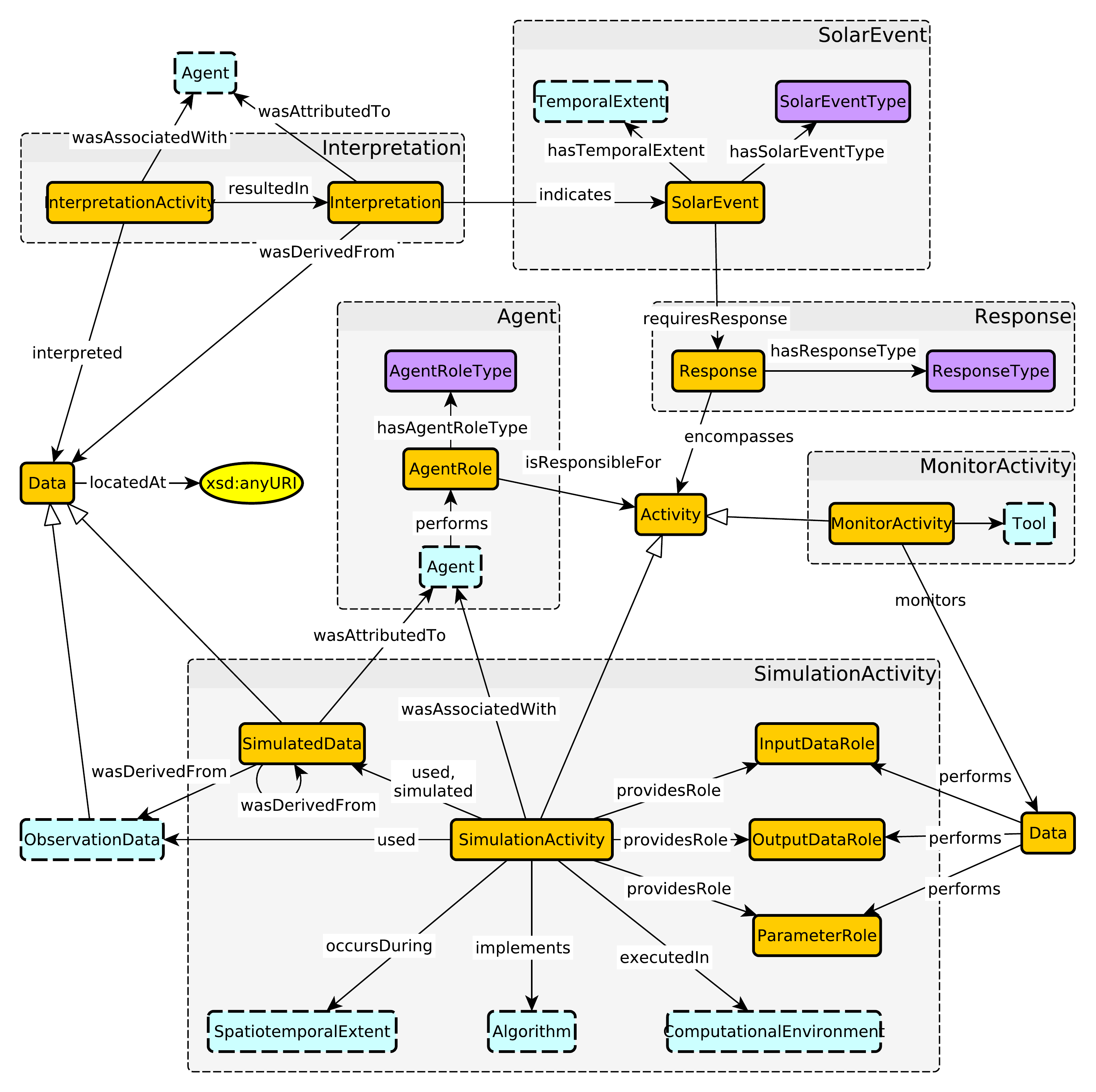}
    \caption{The schema diagram for the \textsf{SimulationActivity} module. The dashed grey boxes denote modules, otherwise utilizes the same notation as previous diagrams.}
    \label{fig:ontology}
\end{figure}

At a top level, the purpose of the ontology is to tie together Activities, such as simulations or interpretations; Agents, such as power grid operators; Solar Events, such as coronal mass ejections; and Responses, such as simulate, monitor, or update. In addition, it is necessary to record the provenance of datasets as they are generated and transformed. We present the entire assembled ontology in Figure \ref{fig:ontology}.

We have previously discussed the pattern used for one Activity: the Simulation Activity. The other activities are Monitor and Interpretation. The Monitor activity is currently left as a stub, which is a tiny pattern for acknowledging further complexity \cite{modl}. The Interpretation Activity is used when an Agent, either a person or computer algorithm, perhaps, interprets some data to indicate that a Solar Event, or not, has a occurred. 

The Agent Role module is based on the Agent Role pattern, equipped with Explicit Typing for the categories of roles. This set of roles is a controlled vocabulary (CV) and would be particular to the domain. In the next section, we discuss how these CVs can be used to add structured domain knowledge.

Agents are responsible for certain Responses. In the same manner, there are different categories of Responses. Responses also encompass an activity. For example, a response to some solar event might be to simulate its impact, and the Activity it encompasses would be exactly that execution.

\subsection{Adding Domain Knowledge with Existential Rules}
\label{ssec:rules}
Although the precise specification that an ontology provides is very helpful for modeling a complex scenario, it is also occasionally difficult to visualize exactly which connections are purely logical and which are essential for a user interacting with the ontology. 
In order to provide a, perhaps more intuitive, representation we show a First-Order Predicate Logic (FOPL) Rule statement that corresponds to entailments of some sample data that would comprise the ontology.\footnote{The abbreviations are as follows: NOAA is the National Oceanic and Atmospheric Administration, SPWC is the Space Weather Prediction Center, and GOES is Geostationary Operational Environment Satellites.} As we are writing the rule in FOPL, we may omit many of the description logic classes that are conceptually important, but play a diminished role in the semantics for a domain expert, and instead focus on the connecting predicates that form the relevant entailment. In this example, we show how a \textsf{SolarEvent}, such as a ``Solar Flare'', leads to a definite chain of \textsf{Responses}. Every solar event defined in the ontology must be represented in this way. This means that if another solar event is known by the ontology, a well-defined sequence of responses to that event must also exist.
\begin{align*}
    & hasSolarEventType(\text{solarevent,``Solar Flare"})  \to \\
    & \exists \text{sequentialresponse,response1,response2,response3,response4,response5,} \\
    & \text{activity1,activity2,activity3,activity4,activity5} \\
    &requiresResponse(\text{solarevent,sequentialresponse}) \wedge \\
    &hasFirstResponse(\text{sequentialresponse,response1}) \wedge \\
    &\qquad encompasses(\text{response1,activity1}) \wedge\\
    & hasResponseType(\text{response1}, \\
    &\qquad \text{``Simulate radiation conditions at Earth using historical analog data"}) \wedge \\
    &hasNextResponse(\text{response1,response2}) \wedge encompasses(\text{response2,activity2}) \wedge \\    
    &hasResponseType(\text{response2}, \\
    &\qquad \text{``Update radiation conditions using NOAA, SWPC,} \\
    &\qquad \text{  and GOES satellites observations"}) \wedge \\
    &hasNextResponse(\text{response2,response3}) \wedge encompasses(\text{response3,activity3}) \wedge \\    
    &hasResponseType(\text{response3}, \\
    &\qquad \text{``Based on radiation conditions run ionospheric models"}) \wedge \\
    &hasNextResponse(\text{response3,response4}) \wedge encompasses(\text{response4,activity4}) \wedge \\    
    &hasResponseType(\text{response4}, \\
    &\qquad \text{``Based on ionospheric currents from ionospheric simulations,} \\
    &\qquad \text{  simulate geomagnetic field"}) \wedge \\
    &hasNextResponse(\text{response4,response5}) \wedge encompasses(\text{response5,activity5}) \wedge \\    
    &hasResponseType(\text{response5}, \\
    &\qquad \text{``Based on geomagnetic field from simulations,} \\
    &\qquad \text{  simulate geomagnetically induced currents"})
\end{align*}

\subsection{Example Triple Data}
Below we reproduce a fragment of what the populated ontology would look like, corresponding to the rule in the previous section. Some triples are omitted for brevity and clarity, however no new data was created besides instances and strings used to fill out the ontology, so this can be viewed as a correct representation.
\begin{verbatim}
:Activity rdf:type owl:Class .
:Response rdf:type owl:Class .
:ResponseType rdf:type owl:Class .
:SequentialResponse rdf:type owl:Class .
:SolarEvent rdf:type owl:Class .
:SolarEventType rdf:type owl:Class .
:encompasses rdf:type owl:ObjectProperty .
:hasFirstResponse rdf:type owl:ObjectProperty .
:hasNextResponse rdf:type owl:ObjectProperty .
:hasResponseType rdf:type owl:ObjectProperty .
:hasSolarEventType rdf:type owl:ObjectProperty .
:requiresResponse rdf:type owl:ObjectProperty .
:asString rdf:type owl:DatatypeProperty .
\end{verbatim}
\begin{verbatim}
:solarEvent rdf:type           owl:NamedIndividual , 
                               :SolarEvent ;
            :hasSolarEventType :solarEventType ;
            :requiresResponse  :sequentialResponse .
:solarEventType rdf:type owl:NamedIndividual ,
                         :SolarEventType ;
                :asString "Solar Flare" .
                
:sequentialResponse rdf:type owl:NamedIndividual , 
                            :SequentialResponse ;
                    :hasFirstResponse :response1 .
                    
:response1 rdf:type owl:NamedIndividual ,
                    :Response ;
           :encompasses :activity1 ;
           :hasNextResponse :response2 ;
           :hasResponseType :responseType1 .
:response2 rdf:type owl:NamedIndividual ,
                    :Response ;
           :encompasses :activity2 ;
           :hasNextResponse :response3 ;
           :hasResponseType :responseType2 .
:response3 rdf:type owl:NamedIndividual ,
                    :Response ;
           :encompasses :activity3 ;
           :hasNextResponse :response4 ;
           :hasResponseType :responseType3 .
:response4 rdf:type owl:NamedIndividual ,
                    :Response ;
           :encompasses :activity4 ;
           :hasNextResponse :response5 ;
           :hasResponseType :responseType4 .
:response5 rdf:type owl:NamedIndividual ,
                    :Response ;
           :encompasses :activity5 ;
           :hasResponseType :responseType5 .
           
:responseType1 rdf:type owl:NamedIndividual ,
                        :ResponseType ;
               :asString "Simulate the radiation conditions 
                          at Earth using historical analog data" .
:responseType2 rdf:type owl:NamedIndividual ,
                        :ResponseType ;
               :asString "Update radiation conditions using NOAA
                          SWPC GOES satellites observations" .
:responseType3 rdf:type owl:NamedIndividual ,
                        :ResponseType ;
               :asString "Based on radiation conditions run 
                          ionospheric models" .
:responseType4 rdf:type owl:NamedIndividual ,
                        :ResponseType ;
               :asString "Based on ionospheric currents from
                          ionospheric simulations, simulate 
                          geomagnetic field" .
:responseType5 rdf:type owl:NamedIndividual ,
                        :ResponseType ;
               :asString "Based on geomagnetic field from
                          simulations, simulate geomagnetically
                          induced currents" .

:activity1 rdf:type owl:NamedIndividual ,
                    :Activity .
:activity2 rdf:type owl:NamedIndividual ,
                    :Activity .
:activity3 rdf:type owl:NamedIndividual ,
                    :Activity .
:activity4 rdf:type owl:NamedIndividual ,
                    :Activity .
:activity5 rdf:type owl:NamedIndividual ,
                    :Activity .
\end{verbatim}

%% file: sections/conclusion.tex
\section{Conclusion}
\label{sec:conc}
This paper has introduced a modular ontology for supporting space weather research, especially in understanding the interactions between agencies and their responsibilities, as well as tracking the usage and provenance of data through observations and simulations. In doing so, we presented a novel Numerical Transformation pattern, which is further modified to fit our use-case via template-based instantiation to produce the SimulationActivity module. This is one of the core patterns that constitute the entire modular ontology. We then go to show how existential rules, making use of instance data in controlled vocabularies, can be used to add and frame domain knowlege. Finally, we discuss a worked example. 

The next steps for this work will be the further development of the modules left as stubs, namely Observation Data, which has obvious connections to the SSN/SOSA\footnote{See \url{https://www.w3.org/TR/vocab-ssn/}.} and the Monitoring activity. We will also be exploring deployment and integration of real world data.


%% file: main.bbl
\begin{thebibliography}{10}
\providecommand{\url}[1]{\texttt{#1}}
\providecommand{\urlprefix}{URL }
\providecommand{\doi}[1]{https://doi.org/#1}

\bibitem{template}
Hammar, K., Presutti, V.: Template-based content {ODP} instantiation. In:
  Hammar, K., Hitzler, P., Krisnadhi, A., Lawrynowicz, A., Nuzzolese, A.G.,
  Solanki, M. (eds.) Advances in Ontology Design and Patterns [revised and
  extended versions of the papers presented at the 7th edition of the Workshop
  on Ontology and Semantic Web Patterns, WOP@ISWC 2016, Kobe, Japan, 18th
  October 2016]. Studies on the Semantic Web, vol.~32, pp. 1--13. {IOS} Press
  (2016). \doi{10.3233/978-1-61499-826-6-1},
  \url{https://doi.org/10.3233/978-1-61499-826-6-1}

\bibitem{opla-cp}
Hirt, Q., Shimizu, C., Hitzler, P.: Extensions to the ontology design pattern
  representation language. In: Janowicz, K., Krisnadhi, A.A.,
  Poveda{-}Villal{\'{o}}n, M., Hammar, K., Shimizu, C. (eds.) Proceedings of
  the 10th Workshop on Ontology Design and Patterns {(WOP} 2019) co-located
  with 18th International Semantic Web Conference {(ISWC} 2019), Auckland, New
  Zealand, October 27, 2019. {CEUR} Workshop Proceedings, vol.~2459, pp.
  76--75. CEUR-WS.org (2019), \url{http://ceur-ws.org/Vol-2459/short2.pdf}

\bibitem{opla}
Hitzler, P., Gangemi, A., Janowicz, K., Krisnadhi, A.A., Presutti, V.: Towards
  a simple but useful ontology design pattern representation language. In:
  Blomqvist, E., Corcho, {\'{O}}., Horridge, M., Carral, D., Hoekstra, R.
  (eds.) Proceedings of the 8th Workshop on Ontology Design and Patterns {(WOP}
  2017) co-located with the 16th International Semantic Web Conference {(ISWC}
  2017), Vienna, Austria, October 21, 2017. {CEUR} Workshop Proceedings,
  vol.~2043. CEUR-WS.org (2017), \url{http://ceur-ws.org/Vol-2043/paper-09.pdf}

\bibitem{McGranaghan_2017}
McGranaghan, R.M., Bhatt, A., Matsuo, T., Mannucci, A.J., Semeter, J.L.,
  Datta-Barua, S.: Ushering in a new frontier in geospace through data science.
  Journal of Geophysical Research: Space Physics  \textbf{122}(12),
  12,586--12,590 (2017). \doi{10.1002/2017JA024835},
  \url{https://agupubs.onlinelibrary.wiley.com/doi/abs/10.1002/2017JA024835}

\bibitem{Narock_2012}
Narock, T., Fox, P.: From science to e-science to semantic e-science: A
  heliophysics case study. Computers \& Geosciences  \textbf{46},  248 -- 254
  (2012). \doi{https://doi.org/10.1016/j.cageo.2011.11.018},
  \url{http://www.sciencedirect.com/science/article/pii/S0098300411004080}

\bibitem{Pulkkinen_2017}
Pulkkinen, A., Bernabeu, E., Thomson, A., Viljanen, A., Pirjola, R., Boteler,
  D., Eichner, J., Cilliers, P.J., Welling, D., Savani, N.P., Weigel, R.S.,
  Love, J.J., Balch, C., Ngwira, C.M., Crowley, G., Schultz, A., Kataoka, R.,
  Anderson, B., Fugate, D., Simpson, J.J., MacAlester, M.: Geomagnetically
  induced currents: Science, engineering, and applications readiness. Space
  Weather  \textbf{15}(7),  828--856 (2017). \doi{10.1002/2016SW001501},
  \url{https://agupubs.onlinelibrary.wiley.com/doi/abs/10.1002/2016SW001501}

\bibitem{Schrijver_2015}
Schrijver, C.J., Kauristie, K., Aylward, A.D., Denardini, C.M., Gibson, S.E.,
  Glover, A., Gopalswamy, N., Grande, M., Hapgood, M., Heynderickx, D.,
  Jakowski, N., Kalegaev, V.V., Lapenta, G., Linker, J.A., Liu, S., Mandrini,
  C.H., Mann, I.R., Nagatsuma, T., Nandy, D., Obara, T., {Paul O’Brien}, T.,
  Onsager, T., Opgenoorth, H.J., Terkildsen, M., Valladares, C.E., Vilmer, N.:
  {{Understanding space weather to shield society: A global road map for
  2015–2025 commissioned by COSPAR and ILWS}}. Advances in Space Research
  \textbf{55}(12),  2745 -- 2807 (2015).
  \doi{https://doi.org/10.1016/j.asr.2015.03.023},
  \url{http://www.sciencedirect.com/science/article/pii/S0273117715002252}

\bibitem{comodide}
Shimizu, C., Hammar, K., Hitzler, P.: Modular graphical ontology engineering
  evaluated. In: Harth, A., Kirrane, S., Ngomo, A.N., Paulheim, H., Rula, A.,
  Gentile, A.L., Haase, P., Cochez, M. (eds.) The Semantic Web - 17th
  International Conference, {ESWC} 2020, Heraklion, Crete, Greece, May 31-June
  4, 2020, Proceedings. Lecture Notes in Computer Science, vol. 12123, pp.
  20--35. Springer (2020). \doi{10.1007/978-3-030-49461-2\_2},
  \url{https://doi.org/10.1007/978-3-030-49461-2\_2}

\bibitem{oplatool}
Shimizu, C., Hirt, Q., Hitzler, P.: A prot{\'{e}}g{\'{e}} plug-in for
  annotating {OWL} ontologies with opla. In: Gangemi, A., Gentile, A.L.,
  Nuzzolese, A.G., Rudolph, S., Maleshkova, M., Paulheim, H., Pan, J.Z., Alam,
  M. (eds.) The Semantic Web: {ESWC} 2018 Satellite Events - {ESWC} 2018
  Satellite Events, Heraklion, Crete, Greece, June 3-7, 2018, Revised Selected
  Papers. Lecture Notes in Computer Science, vol. 11155, pp. 23--27. Springer
  (2018). \doi{10.1007/978-3-319-98192-5\_5},
  \url{https://doi.org/10.1007/978-3-319-98192-5\_5}

\bibitem{modl}
Shimizu, C., Hirt, Q., Hitzler, P.: {MODL:} {A} modular ontology design
  library. In: Janowicz, K., Krisnadhi, A.A., Poveda{-}Villal{\'{o}}n, M.,
  Hammar, K., Shimizu, C. (eds.) Proceedings of the 10th Workshop on Ontology
  Design and Patterns {(WOP} 2019) co-located with 18th International Semantic
  Web Conference {(ISWC} 2019), Auckland, New Zealand, October 27, 2019. {CEUR}
  Workshop Proceedings, vol.~2459, pp. 47--58. CEUR-WS.org (2019),
  \url{http://ceur-ws.org/Vol-2459/paper4.pdf}

\bibitem{moe-chap}
Shimizu, C., Krisnadhi, A., Hitzler, P.: Modular ontology modeling: a tutorial.
  In: Cota, G., Daquino, M., Pozzato, G.L. (eds.) Applications and Practices in
  Ontology Design, Extraction, and Reasoning. Studies on the Semantic Web,
  {IOS} Press (2020), in press

\end{thebibliography}
